\documentclass[sigconf]{acmart}
\usepackage{siunitx}
\usepackage{xcolor,soul}
\usepackage{enumitem}
\usepackage{multirow}

\AtBeginDocument{%
  \providecommand\BibTeX{{%
    \normalfont B\kern-0.5em{\scshape i\kern-0.25em b}\kern-0.8em\TeX}}}

\setcopyright{acmcopyright}
\copyrightyear{2021}
\acmYear{2021}
\acmDOI{10.1145/1122445.1122456}

\acmConference[EarComp '21]{2nd International Workshop on Earable Computing}{Sep 25, 2021}{Online}
\acmBooktitle{EarComp '21: 2nd International Workshop on Earable Computing,
  Sep 25, 2021, online}
\acmPrice{15.00}
\acmISBN{978-1-4503-XXXX-X/21/09}

\begin{document}

\title{Earables for Detection of Bruxism: a Feasibility Study}


\author{Erika Bondareva}
\authornotemark[1]
\affiliation{
  \institution{University of Cambridge}
  \country{United Kingdom}}
 \email{eb729@cam.ac.uk}

\author{Elín Rós Hauksdóttir}
\affiliation{%
  \institution{University of Cambridge}
  \country{United Kingdom}}
\email{erh72@cam.ac.uk}
\authornote{Both authors contributed equally to this research.}

\author{Cecilia Mascolo}
\affiliation{%
  \institution{University of Cambridge}
  \country{United Kingdom}}
  \email{cm542@cam.ac.uk}
  
% \email{cm542@cam.ac.uk}


%%%%%%%%%%%%%%%%%%%%%%%%%%%%%%%%%%% ABSTRACT %%%%%%%%%%%%%%%%%%%%%%%%%%%%%%%%%%%
\begin{abstract}

Bruxism is a disorder characterised by teeth grinding and clenching, and many bruxism sufferers are not aware of this disorder until their dental health professional notices permanent teeth wear. Stress and anxiety are often listed among contributing factors impacting bruxism exacerbation, which may explain why the COVID-19 pandemic gave rise to a bruxism epidemic. It is essential to develop tools allowing for the early diagnosis of bruxism in an unobtrusive manner. This work explores the feasibility of detecting bruxism-related events using earables in a mimicked in-the-wild setting. Using inertial measurement unit for data collection, we utilise traditional machine learning for teeth grinding and clenching detection. We observe superior performance of models based on gyroscope data, achieving an 88\% and 66\% accuracy on grinding and clenching activities, respectively, in a controlled environment, and 76\% and 73\% on grinding and clenching, respectively, in an in-the-wild environment.
\end{abstract}





\maketitle

\section{Introduction}

% spread of wearable technology for health
%\cm{this part of the intro is not needed for this commutnity. go to the point quickly} The first concept of a wearable computer is dated back as long as 1960s, when Thorp and Shannon inventing a cigarette-pack-sized analogue computer for predicting the performance of roulette wheels \cite{jackson-1998}. Since then, the field of wearables has progressed and expanded considerably, with a large variety of wearable devices available commercially, and an even larger variety still in research phase. Nowadays, such devices are used for augmenting gaming experiences, keeping first responders safe by tracking their vital signs, allowing sports enthusiasts and professionals to boost their performance, as well as have a wide application in healthcare sector, from tracking infants' vitals, to detecting falls in elderly users. \cite{wow-chapter} With technological advancements of the past few decades and the rise of personalised medicine, wearables as medical technologies are an especially important area of research and development, offering a potential of low-cost, accessible, and reliable method for early diagnostics and disease progression tracking. \cite{yetisen-2018}
%\cm{I would start the paper here}
% spread of earables and potential of earables for health gathering
In the recent years, wireless earphones with built-in sensors, a.k.a. earables, have been gaining popularity --- earphones are a commodity item providing established functionality, with support for privacy-preserving interaction by allowing the users to access information hands-free in a socially acceptable way, and, most importantly, have unique placement, allowing for numerous applications beyond playing music~\cite{kawsar-2018-esense}. %\cm{citation after the dot?} 
Comparing to other common areas of a human body for wearables placement (e.g. wrist), the ear is significantly more stationary, meaning that the collected signal is less susceptible to the external noise and motion artifacts, while also allowing to capture head and jaw movements, in addition to the activity in the rest of the body~\cite{kawsar-2018-esense}. Due to these unique advantages offered by earable platforms, they are widely explored for health applications. 

% bruxism
Drawing on the ability of wearables to capture jaw movements, we were interested in exploring the feasibility of using earables for detection of a movement disorder, characterised by teeth grinding and clenching, called bruxism. Bruxism affects around 8--13\% of the adult population~\cite{beddis-2018}, and can cause numerous problems, affecting patient's teeth' health, causing headaches and disorders of the temporomandibular joint (TMJ). However, people suffering from bruxism often are unaware of the disorder until it becomes so advanced that the dentist is able to infer the diagnosis from the patient's worn down teeth. The exact cause of bruxism is unknown, but it is believed there is a genetic component to it~\cite{beddis-2018}, and it is usually linked to the levels of stress and anxiety that the patient is experiencing. With the COVID-19 pandemic having lasted for over a year and having had a major impact on the lives of nearly every individual, dentists are warning of another, accompanying, bruxism epidemic~\cite{PandemicAndBruxism:online}, prompting the issue of tooth wear and other side effects of bruxism.

%% how is it currently diagnosed
Existing methods for diagnosing bruxism tend to be unreliable or invasive~\cite{inEarClenching:online}. Most cases that are detected in earlier stages are based on self-reporting: for example, when a sleep partner notices grinding sounds, or when the patient reports TMJ pain. However, previous research indicates that the validity of the self-reported assessment of bruxism is low to modest and therefore is usually not sufficient for diagnosis of bruxism~\cite{validitySelfReport:online}. The golden standard for definitive diagnosis of bruxism is electromyogram (EMG) of the masticatory muscles by polysomnography audio-visual (PSG-AV) recording~\cite{currentMethods:online, BruxismALitratureReview:online, Validity19PSG:online}. This method is performed in a controlled environment, and due to its complexity and high cost PSG-AV is not used for the assessment of bruxism in daily clinical practice.%~\cite{currentMethods:online, BruxismALitratureReview:online, Validity19PSG:online}. 

% scope of this project
To address the necessity of detecting bruxism in a non-invasive and low-cost way, this research presents a methodology to detect teeth clenching and grinding through an earable device by using traditional  machine learning approaches. We collected accelerometer and gyroscope data from 17 participants using eSense wireless earbuds with a built-in inertial measurement unit (IMU), using the data from 13 participants for machine learning. %\eh{data from 17 participants? and add the the data from 13 was used? } 
The data comprised of participants grinding and clenching their teeth in a controlled environment, as well as performing bruxism-mimicking actions while engaging in routine activities that would simulate in-the-wild deployment. Namely, these activities included head movement, listening to music, walking, talking, chewing, and drinking. 

To increase the inference accuracy, we collected the signals from both ears of each participant, and we extracted both time and frequency domain features. After preprocessing the collected data, we used traditional machine learning methods to develop a bruxism detection algorithm, using random forest (RF) and support vector machine (SVM). The detection algorithms were evaluated using both clean data and the more realistic signals collected with participants engaging in routine behaviours that could potentially affect the classification of teeth grinding events.  

We show that by utilising traditional machine learning methods we can detect teeth grinding in a controlled environment with accuracy up to 88\%. Additionally, we can detect teeth grinding with up to 76\% accuracy for datasets mimicking real-world scenarios. The performance on clenching detection task is poorer, although still shows promise, with us achieving up to 73\% accuracy on in-the-wild setting.

% unique contribution outline
The main contributions of this work are as follows:
\begin{itemize}
\item We present a novel dataset compiled as a part of this study, which contains teeth grinding and clenching data collected through earables from 13 participants, in both noise-free environments and while performing routine activities to mimic in-the-wild data.
\item We evaluate traditional signal processing and machine learning techniques for development of teeth grinding and clenching detection algorithms.
\item We show the potential of detecting bruxism through earables by achieving 76\% accuracy on in-the-wild teeth grinding and 73\% accuracy on in-the-wild teeth clenching, and provide ideas for future directions.
\end{itemize}

\section{Related Work}

EMG is a technology commonly utilised for a non-trivial task of bruxism diagnosis, capable of detecting mastication muscle movement. However, according to~\cite{emg-limitations} EMG may detect muscle movement that is not necessarily related to bruxism, limiting the accuracy of methods relying solely on EMG. In addition, accuracy of portable EMG recorders for bruxism detection was reported as being unsatisfactory.~\cite{emg-accuracy} EMG can also be seen as obtrusive, relying on electrodes placed on the face for data collection. Therefore, it is important to explore alternative sensing modalities for detection of bruxism-related events.

There have been numerous efforts in the earables field exploring the potential of in-ear wearables for detection of activities related to the mouth. 

% earables for oral activity
A number of studies looked at detecting jaw and mouth movements by using earables. CanalSense presented a jaw, face, and head movement recognition system based on detecting changes in the air pressure inside the ear canal, using barometers embedded in earphones ~\cite{CanalSen60:online}. Another system, EarSense, sensed teeth gestures by detecting vibrations in the jaw that propagate through the skull to the ear, creating oscillations in the earphone speaker’s diaphragm ~\cite{EarSense48:online}. Other works looked at developing a separate system, rather than utilising earbuds, for detection of jaw movements. One such example would be the Outer Ear Interface that measured the deformation in the ear canal using proximity sensors caused by the lower jaw movements ~\cite{OEI:online}.

There have also been successful attempts at forming a human-computer interaction system based on unvoiced jaw movement tracking. JawSense considered the neurological and anatomical structure of the human jaw and cheek upon system design, and achieved successful classification of nine phonemes based on the muscle deformation and vibrations caused by unvoiced speaking ~\cite{JawSense6:online}.

~\cite{inEarClenching:online} looked at using gyroscope data from an in-ear wearable for jaw clenching, which is an important part of what we set out to achieve in this feasibility study. The reported results had an error rate of 1\% when the participant was seated and 4\% when the participant moved, but the work was based on a single participant, and did not explore the detection of grinding. 

% earables for bruxism
Multiple works explored detection of bruxism using wearable devices, however, most of them are not as inconspicuous as a pair of earbuds. ~\cite{aroundEar:online} introduced comfortable around the ear sensors (cEEGrids) for detecting awake bruxism, analysing bruxism-related events in contrast to the other facial activity events, such as chewing and speaking.  ~\cite{mouthGuard:online} developed a wearable mouthguard with a force sensor to analyse teeth clenching during exercise. ~\cite{SBNaturalEnviorment:online} proposed to detect sleep bruxism by using electromyography (EMG) and electrocardiogram (ECG) signals in combination, which produced substantially better results than using only EMG. ~\cite{forceSensor:online} developed a system consisting of an interrogator/reader and a passive sensor that could be used to record bruxism-related events by placing the system in a dental splint. Finally, ~\cite{biteGuard:online} developed a bite guard designed to analyse bruxism, with the monitoring achieved through a novel pressure-sensitive polymer. 

To the best of our knowledge, this is the first work that looks at using wireless earbuds with built-in IMUs for detection of grinding and clenching with the goal of diagnosing and tracking bruxism. 

%\eh{I think this is not really correct. [16] looks into clenching by using eSense. We could add by using ML to make it a bit more true.. or what}
%\eb{I will reword this, since we look specifically at grinding, which they don't}

\section{Study design}

An in-ear multisensory stereo device eSense was used for data collection. Specifically, we collected three-axis accelerometer and three-axis gyroscope data from the built-in IMU.

To collect the aforementioned data a mobile application, called eSense Client, was used for connecting via Bluetooth to the eSense earbuds and collecting raw IMU data. Due to the COVID-19 pandemic, all of the experiments were carried out remotely. Therefore, for annotating the data collected from the earables, a timestamped video was recorded using Zoom \cite{ZoomSecure:online}. The collected eSense data was labelled by matching it with the video recording's timestamp.

Worth noting, that the eSense earbuds~\cite{kawsar-2018-esense, eSense2:online} contain  IMU only in the left earbud. Therefore, we used two left earbuds from two pairs of eSense for data collection, to explore the variation in accuracy for two sides, as well as potentially use the data gathered from both right and left ears together. Indeed, we discovered that participants typically had a dominant chewing side, which resulted in mastication %\cm{mastication?} 
muscles on one side of the face being greater developed. This resulted in data from one of the ears being more valuable for correct classification of bruxism-related activities, but since the dominant chewing side varies for different people, we had to use data from both right and left ear together.

The study was approved by the ethics committee in the Department of Computer Science and Technology at the University of Cambridge. Informed consent was collected from the study participants. We ensured that no identifiable information was collected, and deleted the videos after the IMU data was labelled. Upon consulting a dental health professional, we also excluded any participants who suffer from bruxism to avoid any further damage to their teeth, as well as compiled a short questionnaire aimed at identifying potential participants who might be unknowingly suffering from bruxism. The questionnaire was based on Shetty's et al. research \cite{BruxismALitratureReview:online}, but amended in collaboration with a certificated dentist to suit the needs of our study. In addition to these precautions, we also included a compulsory set of simple jaw massage exercises typically used in TMJ physiotherapy, aimed at alleviating any tension in the TMJ that the experiment might have inadvertently caused.

% description of the dataset

For the data collection, in total 17 participants were recruited, 12 females and 5 males, with the youngest participant aged 23 and the oldest aged 61. After verifying the quality of the collected data, data from four participants were discarded due to being compromised during data collection. Therefore, data from 13 participants were used for this research. 

The aim of this study was to assess the feasibility of using in-ear wearables for detection of bruxism-related events, specifically teeth grinding and clenching. To address this goal, participants were asked to perform the following seven human-centred sensing experiments (with experiments 1-6 conducted with the participant in a sitting position):
%\eb{what position where the participants in?} \eh{sorry for not being clear enough. Think we should maybe add the the numbers 1-7 to indicate which experiments are number what and then switch Listening to music and walking around. Then 1-6 are performed in a sitting position }

\textbf{1. Control experiment:}
\begin{enumerate}[label=(\alph*),leftmargin=*,align=left]
\item Grind teeth for 5 seconds, pause for 5 seconds, repeat 6 times.
\item Clench teeth for 5 seconds, pause for 5 seconds, repeat 6 times.
\end{enumerate}

\textbf{2. Moving head side to side:}
\begin{enumerate} [label=(\alph*),leftmargin=*,align=left]
\item Look right and left for 30 seconds.
\item Look right and left for the duration of 5 seconds while grinding, pause the grinding for 5 seconds, and repeat 6 times.
\item Look right and left for the duration of 5 seconds while clenching, pause the clenching for 5 seconds, and repeat 6 times.
\end{enumerate}

\textbf{3. Chewing:}
\begin{enumerate} [label=(\alph*),leftmargin=*,align=left]
\item Eat half a slice of bread.
\item Chew gum for 30 seconds.
\end{enumerate}

\textbf{4. Read the provided text out loud for 30 seconds.}

\textbf{5. Drink 250 ml of water.}

\textbf{6. Listening to music:}
\begin{enumerate}[label=(\alph*),leftmargin=*,align=left]
\item Listen to music for 30 seconds.
\item While listening to music, grind teeth for 5 seconds, then pause for 5 seconds, and repeat 6 times.
\item While listening to music, clench teeth for 5 seconds, then pause for 5 seconds, and repeat 6 times.
\end{enumerate}

\textbf{7. Walk around in a quiet room:}
\begin{enumerate}[label=(\alph*),leftmargin=*,align=left]
\item Walk around in a quiet room for 30 seconds.
\item Walk around for the duration of 5 seconds while clenching, pause the clenching for 5 seconds, and repeat 6 times.
\end{enumerate}

The experiment took around 35 minutes for each participant to complete from start to finish.

To the best of our knowledge, this is the most advanced and sophisticated dataset that exists for in-ear IMU data of teeth grinding and clenching. 

The experiments were designed each with a specific purpose in mind. While the first experiment was intended as control in a quiet, albeit unrealistic environment, the rest of the experiments were designed to either recreate activities that are known to interfere with signals collected via earbuds, such as moving the head, walking, as well as listening to music -- an especially important task, keeping in mind that the primary purpose of earbuds is playing music. Other activities, such as chewing, drinking, and reading, were meant to recreate the activities in which a typical user is likely to engage daily, which include a significant involvement of mastication muscles, which might result in a signal similar to either clenching or grinding.

\section{Methodology}

\subsection{Data collection}

IMU (accelerometer and gyroscope) data was collected from both ears of the participants with a sampling rate of \SI{5}{Hz}. This yielded approximately 25 minutes of IMU data per participant, containing 12 columns: X, Y, and Z axes for the acceleration and three axes for the gyroscope data collected from one ear, and the same data collected from the second ear. In addition to the IMU data, we also recorded a video of the participant performing the tasks, to use as ground-truth for correlating the collected IMU data to grinding and clenching events. Specifically, the videos were used to note down the start and end times of each general activity (such as moving head, chewing, walking, etc.), and also to note down the start and end times of grinding and clenching events during these activities. The times with no grinding or clenching events were noted down as silent periods, regardless of the general activity performed.

% windowing and labelling
\subsection{Segmentation and Labelling}

Dealing with data in time domain, segmentation into shorter windows was necessary. We used a sliding window with \SI{1.6}{\second} length and 50\% (or \SI{0.8}{\second}) overlap, to minimise the risk of missing the transition from one event into another.

Creating a label for each window posed a challenge due to the fact that sometimes the window would contain a transition from one event to another, such as transitioning from silent to grinding, making the labelling non-trivial. We explored two labelling methodologies:
\begin{itemize}
    \item labelling the window as the dominant event in that window: if the larger portion of the window contains the grinding data and smaller portion of silent data, the window would be labelled as grinding. If the split is equal, then the window is labelled as silent. %\eb{what happens if the split is equal? 4 samples grind and 4 silent?} \eh{it selects the silent labelle}
    \item only labelling the window as silent if all the samples within the window are silent, and no amount of grinding or clenching event present.
\end{itemize}
Based on the preliminary comparison of the methodologies, it appeared that labelling the window according to the dominant event yielded superior performance, due to which this was the method that we chose for further analysis.

% feature extraction
\subsection{Feature Extraction}

%\eb{check with Elin if rerun experiments with correct feature vectors}

As input for the machine learning algorithms we used a number of various features. Raw signal from both ears was used, resulting in 8 datapoints for each axis for each ear, yielding 48 features. % \eh{not sure what u mean here, what 8 points, I think 6?} 
Then, we used a sum of vector magnitudes ($\text{SOVM}$), which was calculated using the following equation: 

\begin{equation*}
\text{SOVM} = \sqrt{ x_R^2 + y_R^2 + z_R^2 } + \sqrt{ x_L^2 + y_L^2 + z_L^2 }, 
\end{equation*}
where $x$, $y$, and $z$ represent a single value collected from the sensor, corresponding to one of the axes, and $L$ and $R$ representing the signals collected from left and right ear, respectively.

\begin{figure} [b]
    \centering
    \includegraphics[width = 11cm, trim={4.5cm 3.5cm 3cm 3.5cm}, clip]{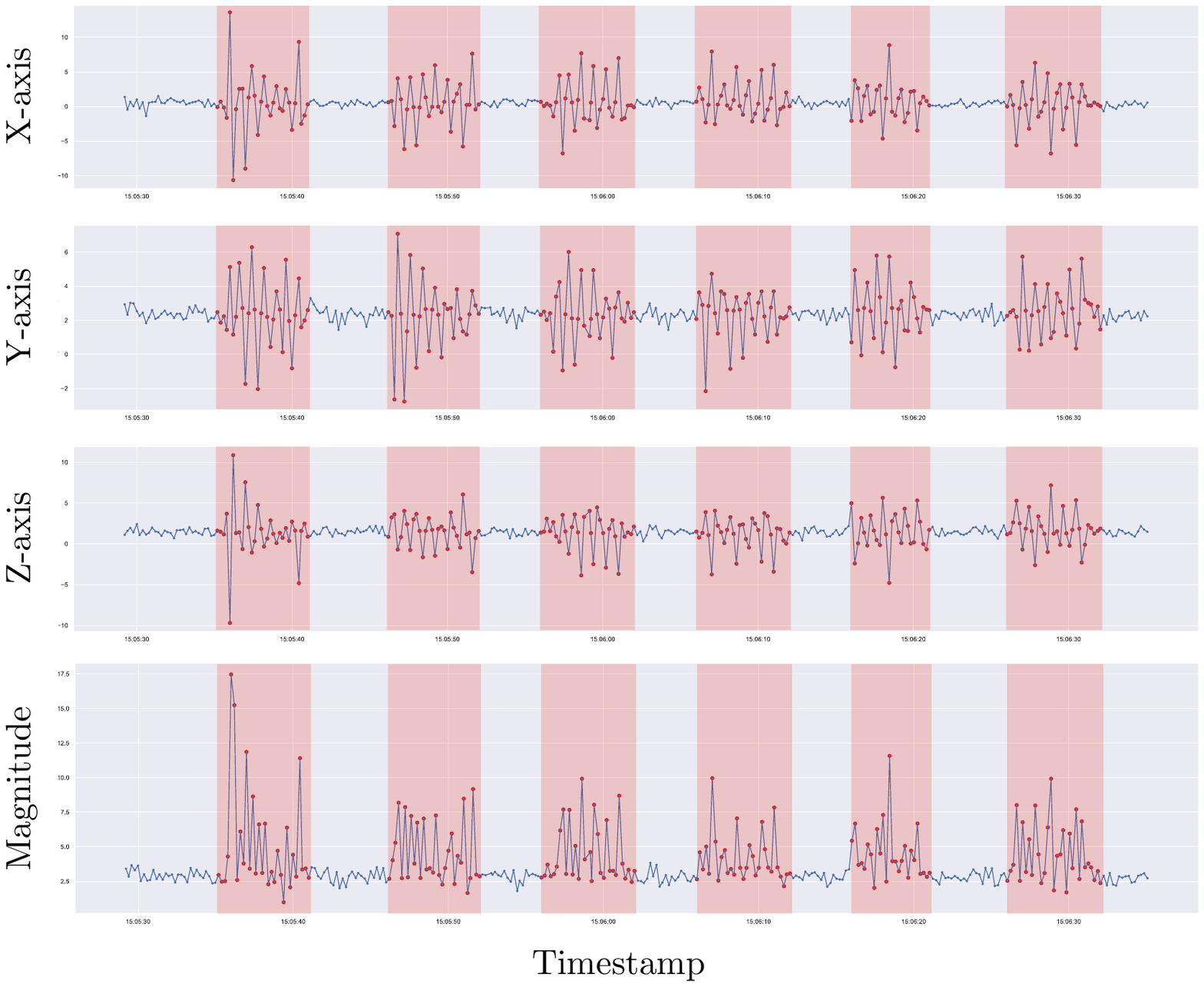}
    \caption{Raw gyroscope data collected with the participant grinding teeth, with the grinding events coloured in red, and periods with no grinding coloured in grey.}
    \label{fig:gyro}
\end{figure}

The raw data collected from the three axes and the calculated SOVM can be seen in Figure~\ref{fig:gyro}.

We utilised commonly-used Python libraries Librosa and Scipy to extract a few additional features: Mel Frequency Cepstral Coefficients (MFCCs), spectral flatness, spectral centroid, and poly features calculated for the sum of vector magnitudes, and a mean was calculated for each of the additional feature vectors. Then, maximum, minimum, mean, standard deviation, and absolute deviation of the signal amplitude was extracted from the sum of vector magnitudes and concatenated with the rest of the features. Finally, zero-crossing rate for each of the axes was averaged and concatenated. This yielded a total of 71 features.

% classification
\subsection{Classification}

Five traditional machine learning classifiers were explored: decision tree (DT),  k-nearest neighbours (k-NN),  logistic regression (LR), random forest (RF), and support vector machine (SVM). In this work, the results for RF and SVM are reported, since they demonstrated the best performance during preliminary evaluation. 

For evaluating the detection algorithm, we used a leave-one-out approach, commonly utilised in machine learning problems with limited number of participants. %\eb{check with Elin that she did a leave-one-out} 
Then, mean and standard deviation were calculated for each of the performance metrics across the resulting values.

To evaluate the algorithm, we compared accuracy, precision, recall, and f1-score. While low number of false positives (FP) and false negatives (FN) is desirable, it is important to keep in mind that in this scenario occasional FP or FN might be acceptable due to the ability to infer the overall diagnosis only if multiple bruxism-related events are detected.

% evaluation method
\subsection{Experiment design}

The goal of this study was to assess the feasibility of detecting bruxism-related events in a controlled and a mimicked in-the-wild environment through in-ear wearables using IMU signals. This goal informed our experiment design. In this paper we present the performance of two traditional machine learning algorithms, RF and SVM, on data collected from two different sensors, accelerometer and gyroscope, for the following tasks:

\textbf{Task 1}: detection of bruxism-related events in a controlled environment, with minimal external noise and no other actions performed, participant being still:
    \begin{enumerate}[label=(\alph*)]
        \item for teeth grinding (580:388 windows of grinding:silent data);
        \item for teeth clenching (589:438 windows of clenching:silent data).
    \end{enumerate}
    
\textbf{Task 2}: detection of bruxism-related events in a mimicked in-the-wild environment (in order to evaluate algorithm performance in a more realistic setting), with participants performing a range of routine activities:
    \begin{enumerate}[label=(\alph*)]
        \item teeth grinding and clenching during no general activity, while moving head, and while listening to music (1658:2334 grinding:silent and 1695:2334 clenching:silent windows);
        \item activities from task 2b with the addition of other routine activities, such as chewing bread, chewing gum, reading out loud (to imitate speaking), drinking water, and walking. Worth noting, that the additional activities do not include grinding or clenching events, and are intended for testing the detection algorithm performance on actions that are known to either be relatively noisy or involve significant jaw activity that may be misclassified as teeth grinding or clenching (1658:4825 grinding:silent and 1695:4815 clenching:silent windows).
    \end{enumerate}

\section{Results and Discussion}

\subsection{Bruxism Detection in Controlled Environment}

Task 1 was designed with the purpose of assessing the feasibility of detecting teeth grinding and clenching using earbuds with a built-in IMU in a controlled environment, with the participant completely still. Based on leave-one-out validation, comparing SVM to RF demonstrates that SVM performs better on both grinding and clenching detection. It is obvious that using gyroscope data yields a significantly higher performance, achieving 89\% and 60\% on grinding and clenching events detection, respectively. For the controlled experiment, RF achieves superior performance only on detection of grinding events using accelerometer data (70\% accuracy), in comparison to detection of grinding events using accelerometer and SVM (66\% accuracy). However, both results are significantly worse than the detection accuracy achieved using gyroscope data. We also calculated precision, recall, and f-1 score, and these metrics as well as the standard deviation for the leave-one-out cross-validation are reported in Table~\ref{table:task1}. 

\begin{table}[b]
\caption{Detection of grinding (denoted as \textbf{Gr.}) and clenching (denoted as \textbf{Cl.}) by SVM and RF algorithms on Task 1. The values reported are mean$\pm$stdev.}
\label{table:task1}
\begin{tabular}{clcccc}
\toprule
                     &           & \multicolumn{2}{c}{\textbf{Gyroscope}} & \multicolumn{2}{c}{\textbf{Accelerometer}}                             \\
                     &           & SVM           & RF            & SVM                           & RF                            \\
\midrule
\multirow{4}{*}{\textbf{Gr.}} & Accuracy  & 0.88±0.05     & 0.85±0.08     & 0.66±0.12                     & 0.70±0.11                     \\
                     & Precision & 0.89±0.04     & 0.87±0.07     & 0.69±0.19                     & 0.70±0.16                     \\
                     & Recall    & 0.88±0.05     & 0.84±0.09     & 0.63±0.12                     & 0.67±0.12                     \\
                     & f1-score  & 0.88±0.05     & 0.83±0.10     & 0.57±0.17                     & 0.63±0.15                     \\
\midrule
\multirow{4}{*}{\textbf{Cl.}} & Accuracy & 0.60±0.06 & 0.57±0.06 & \multicolumn{1}{l}{0.58±0.05} & \multicolumn{1}{l}{0.55±0.06} \\
                     & Precision & 0.56±0.13     & 0.54±0.10     & \multicolumn{1}{l}{0.40±0.15} & \multicolumn{1}{l}{0.53±0.12} \\
                     & Recall    & 0.57±0.06     & 0.54±0.07     & \multicolumn{1}{l}{0.51±0.02} & \multicolumn{1}{l}{0.52±0.08} \\
                     & f1-score  & 0.52±0.11     & 0.52±0.11     & \multicolumn{1}{l}{0.40±0.06} & \multicolumn{1}{l}{0.49±0.08} \\
\bottomrule
\end{tabular}
\end{table}

\subsection{Bruxism Detection In-The-Wild}

We demonstrated that in-ear wearables show promise for detection of bruxism-related events in a controlled environment. But, naturally, it is important to also analyse whether earables could offer a viable solution for unobtrusive bruxism detection in-the-wild. For this purpose, Task2(a) and Task2(b) were designed to test the bruxism-related activity detection approach while the study participants were performing other actions. Specifically, Task2(a) focused on teeth grinding and clenching in silence, as well as these events while listening to music or moving the head. Task2(b) presented an even more complex problem, adding other activities that involve substantial jaw movement.

For Task2(a), SVM model on gyroscope data showed the best performance, for teeth grinding detection yielding 73\% accuracy, and reaching 61\% accuracy for teeth clenching detection. Task2(b) proved to be a more challenging experiment, which can be observed from reduced performance on precision, recall, and f-1 score. However, RF still performs sufficiently well on gyroscope data, yielding a 68\% and 61\% precision on grinding and clenching, respectively. SVM was incapable of detecting clenching using accelerometer data, predicting that all the testing data do not contain clenching instances. Therefore, no metrics are reported for this algorithm. In general, acceleration-based performance was insufficient, proving that gyroscope provides more valuable data for bruxism detection. Detailed results for these experiments can be seen in Tables~\ref{Table:task2a} and~\ref{Table:task2b}.

\begin{table}[]
\caption{Performance on detection of teeth grinding and clenching on Task2(a).}
\label{Table:task2a}
\begin{tabular}{llcccc}
\toprule
                     &           & \multicolumn{2}{c}{\textbf{Gyroscope}} & \multicolumn{2}{c}{\textbf{Accelerometer}}                             \\
                     &           & SVM           & RF            & SVM                           & RF                            \\
                     \midrule
\multirow{4}{*}{\textbf{Gr.}} & Accuracy  & 0.73±0.05     & 0.73±0.08     & 0.55±0.04                     & 0.56±0.07                     \\
                     & Precision & 0.74±0.06     & 0.73±0.08     & 0.43±0.10                     & 0.51±0.12                     \\
                     & Recall    & 0.70±0.05     & 0.71±0.08     & 0.49±0.03                     & 0.52±0.07                     \\
                     & f1-score  & 0.70±0.06     & 0.71±0.09     & 0.42±0.06                     & 0.48±0.09                     \\
\midrule
\multirow{4}{*}{\textbf{Cl.}} & Accuracy & 0.61±0.05 & 0.60±0.04 & \multicolumn{1}{l}{0.57±0.03} & \multicolumn{1}{l}{0.52±0.06} \\
                     & Precision & 0.60±0.08     & 0.57±0.07     & \multicolumn{1}{l}{0.36±0.15} & \multicolumn{1}{l}{0.45±0.08} \\
                     & Recall    & 0.56±0.06     & 0.57±0.05     & \multicolumn{1}{l}{0.50±0.01} & \multicolumn{1}{l}{0.49±0.02} \\
                     & f1-score  & 0.52±0.08     & 0.55±0.07     & \multicolumn{1}{l}{0.38±0.04} & \multicolumn{1}{l}{0.42±0.04} \\
                     \bottomrule
\end{tabular}
\end{table}

\begin{table}[]
\caption{Performance on detection of teeth grinding and clenching on Task2(b).}
\label{Table:task2b}
\begin{tabular}{llcccc}
\toprule
                     &           & \multicolumn{2}{c}{\textbf{Gyroscope}} & \multicolumn{2}{c}{\textbf{Accelerometer}}                    \\
                     &           & SVM           & RF            & SVM                  & RF                            \\
                     \midrule
\multirow{4}{*}{\textbf{Gr.}} & Accuracy  & 0.74±0.03     & 0.76±0.05     & 0.73±0.03            & 0.67±0.11                     \\
                     & Precision & 0.53±0.18     & 0.68±0.07     & 0.46±0.14            & 0.48±0.10                     \\
                     & Recall    & 0.51±0.02     & 0.61±0.06     & 0.50±0.02            & 0.49±0.05                     \\
                     & f1-score  & 0.45±0.04     & 0.61±0.07     & 0.44±0.04            & 0.45±0.07                     \\
                     \midrule
\multirow{4}{*}{\textbf{Cl.}} & Accuracy & 0.74±0.02 & 0.73±0.02 & \multicolumn{1}{c}{\multirow{4}{*}{N/A}} & \multicolumn{1}{l}{0.70±0.04} \\
                     & Precision & 0.39±0.05     & 0.61±0.05     & \multicolumn{1}{l}{} & \multicolumn{1}{l}{0.41±0.05} \\
                     & Recall    & 0.50±0.00     & 0.54±0.02     & \multicolumn{1}{l}{} & \multicolumn{1}{l}{0.49±0.02} \\
                     & f1-score  & 0.43±0.01     & 0.52±0.03     & \multicolumn{1}{l}{} & \multicolumn{1}{l}{0.44±0.03}\\
                     \bottomrule
\end{tabular}
\end{table}

 We also evaluated which tasks have the most impact on detection of bruxism, concluding that head movements and reading out loud have the most impact on correct classification of grinding events, and head movements, walking, and drinking have the most impact on clenching detection.

\subsection{Limitations and Future Work}

Although this work is intended as a feasibility study to assess the capability of in-ear wearables to detect bruxism-related events, it is nevertheless important to highlight its limitations  and discuss some potential avenues worth exploring in the future.

% people are typically sleeping when gridning, unconcious

Teeth grinding, which is the typical symptom of bruxism, usually happens when the person is not fully conscious, most often exhibited in patients' sleep. Our dataset consists of people who are unlikely to be actually suffering from bruxism, and the data is collected with the participants being awake and fully conscious, which means that the data which would be collected from real bruxism sufferers might be slightly different. However, earbuds for sleep have started to appear on the market, the sole purpose of which is to provide comfortable noise cancellation, which is very promising for unobtrusive detection of bruxism during sleep.
%\cm{I would mention that earables which aid sleep are starting to be produced and they would enable us to gather data also during sleep: if you google in ear warables for sleep you find some}

% clenching performancce is lower than grinding

% the windowing approach is not the most acceptable for clenching, especially not with accelerometer, since the clenching would only be detectable when the clenching is initiated, not when clenching is in progress. Gyroscope might be better, but still not ideal. Better labelling needs to be considered, potentially a RNN that would allow processing of a time series data

Our results show that the accuracy of clenching detection is lower than that of grinding. However, it is important to note that grinding is an action that involves continuous movement in the jaw joint, resulting in continuously changing IMU data. Clenching, on the other hand, only involves changes in acceleration and angular velocity when the action is initiated, i.e. the person clasps their teeth together, and not when the action is in progress, i.e. the teeth are clasped and the jaw is not moving. This may explain the poorer performance on detection of clenching events. Going forward, exploring alternative ways of data segmentation and labelling might be useful, as well as more advanced machine learning approaches that are capable of dealing with time-series data, such as recurrent neural networks (RNNs). 

From the classification perspective, given recent advances in deep learning, it would be imperative to explore the potential of deep learning on raw IMU data from earables for detection of bruxism-related events.

% could explore using gyro and accel together, as well as in-ear audio

Other potential areas of research would include exploring sensor fusion and considering accelerometer and gyroscope data in combination, as well as potentially investigating the feasibility of using audio collected with a microphone pointing inside the ear~\cite{dong-2021}.%\cm{cite the mobisys of Dong?}

% should explore more advanced ML algorithms, e.g. deep learning

% by considering more complext machine learning algorithms, will need more data, therefore could look into bruxism data augmentation

Finally, given the complexity of collecting and labelling bruxism data, it would be interesting to explore IMU data augmentation, as well as potentially generating synthetic data. 

%\eh{ 71 Features total and no PCA applied}
\balance
\section{Conclusions}

This work presents a feasibility study on detection of bruxism-related events using earables. During a bruxism epidemic, which was potentially exacerbated by the COVID-19 pandemic, it is essential to devise a low-cost, unobtrusive, and socially acceptable method for bruxism detection, which would allow to diagnose the patients in early stages of the disorder, before irreversible tooth wear occurs. We compiled a first extensive dataset of teeth grinding and clenching data collected via earbuds with a built-in IMU. In addition to collecting these data in a controlled environment, we also collected data mimicking in-the-wild signal by asking the users to simulate teeth grinding and clenching while performing other activities, or to engage in routine activities that require substantial jaw involvement. 

By using traditional machine learning methods, we concluded that SVM and RF yield the best performance. Gyroscope data appears to be much more valuable than acceleration data for identification of bruxism-related events. We achieved 88\% and 66\% accuracy on teeth grinding and clenching, respectively, in a controlled environment. We also demonstrate the potential of this technology in a mimicked in-the-wild environment, achieving 76\% and 73\% accuracy on teeth grinding and clenching, respectively.


\begin{acks}
This work was supported by the UK Engineering and Physical Sciences Research Council (EPSRC) grant EP/L015889/1 for the Centre for Doctoral Training in Sensor Technologies and Applications, ERC  Project 833296 (EAR), and Nokia Bell Labs through their
donation for the Centre of Mobile, Wearable Systems and Augmented Intelligence.
\end{acks}
\balance
\bibliographystyle{ACM-Reference-Format}
\bibliography{main}


\begin{thebibliography}{22}

%%% the Web address.

\ifx \showCODEN    \undefined \def \showCODEN     #1{\unskip}     \fi
\ifx \showDOI      \undefined \def \showDOI       #1{#1}\fi
\ifx \showISBNx    \undefined \def \showISBNx     #1{\unskip}     \fi
\ifx \showISBNxiii \undefined \def \showISBNxiii  #1{\unskip}     \fi
\ifx \showISSN     \undefined \def \showISSN      #1{\unskip}     \fi
\ifx \showLCCN     \undefined \def \showLCCN      #1{\unskip}     \fi
\ifx \shownote     \undefined \def \shownote      #1{#1}          \fi
\ifx \showarticletitle \undefined \def \showarticletitle #1{#1}   \fi
\ifx \showURL      \undefined \def \showURL       {\relax}        \fi
% invisible to TeX
\providecommand\bibfield[2]{#2}
\providecommand\bibinfo[2]{#2}
\providecommand\natexlab[1]{#1}
\providecommand\showeprint[2][]{arXiv:#2}

\bibitem[\protect\citeauthoryear{Ando, Kubo, Shizuki, and Takahashi}{Ando
  et~al\mbox{.}}{2017}]%
        {CanalSen60:online}
\bibfield{author}{\bibinfo{person}{Toshiyuki Ando}, \bibinfo{person}{Yuki
  Kubo}, \bibinfo{person}{Buntarou Shizuki}, {and} \bibinfo{person}{Shin
  Takahashi}.} \bibinfo{year}{2017}\natexlab{}.
\newblock \showarticletitle{CanalSense: Face-Related Movement Recognition
  System Based on Sensing Air Pressure in Ear Canals}. In
  \bibinfo{booktitle}{\emph{Proceedings of the 30th Annual ACM Symposium on
  User Interface Software and Technology}} (Qu\'{e}bec City, QC, Canada)
  \emph{(\bibinfo{series}{UIST '17})}. \bibinfo{publisher}{Association for
  Computing Machinery}, \bibinfo{pages}{679–689}.
\newblock
\urldef\tempurl%
\url{https://doi.org/10.1145/3126594.3126649}
\showDOI{\tempurl}


\bibitem[\protect\citeauthoryear{Beddis, Pemberton, and Davies}{Beddis
  et~al\mbox{.}}{2018}]%
        {beddis-2018}
\bibfield{author}{\bibinfo{person}{H. Beddis}, \bibinfo{person}{M. Pemberton},
  {and} \bibinfo{person}{Stephen Davies}.} \bibinfo{year}{2018}\natexlab{}.
\newblock \showarticletitle{Sleep Bruxism: an Overview for Clinicians}.
\newblock \bibinfo{journal}{\emph{British Dental Journal}}
  \bibinfo{volume}{225}, \bibinfo{number}{6} (\bibinfo{year}{2018}),
  \bibinfo{pages}{497--501}.
\newblock
\showISBNx{1476-5373}
\urldef\tempurl%
\url{https://doi.org/10.1038/sj.bdj.2018.757}
\showDOI{\tempurl}


\bibitem[\protect\citeauthoryear{Bedri, Byrd, Presti, Sahni, Gue, and
  Starner}{Bedri et~al\mbox{.}}{2015}]%
        {OEI:online}
\bibfield{author}{\bibinfo{person}{Abdelkareem Bedri}, \bibinfo{person}{David
  Byrd}, \bibinfo{person}{Peter Presti}, \bibinfo{person}{Himanshu Sahni},
  \bibinfo{person}{Zehua Gue}, {and} \bibinfo{person}{Thad Starner}.}
  \bibinfo{year}{2015}\natexlab{}.
\newblock \showarticletitle{Stick It in Your Ear: Building an in-Ear Jaw
  Movement Sensor} \emph{(\bibinfo{series}{UbiComp/ISWC'15 Adjunct})}.
  \bibinfo{publisher}{Association for Computing Machinery},
  \bibinfo{pages}{1333–1338}.
\newblock
\urldef\tempurl%
\url{https://doi.org/10.1145/2800835.2807933}
\showDOI{\tempurl}


\bibitem[\protect\citeauthoryear{Casett, R{\'e}us, Stuginski-Barbosa,
  Porporatti, Carra, Peres, Canto, and Manfredini}{Casett
  et~al\mbox{.}}{2017}]%
        {Validity19PSG:online}
\bibfield{author}{\bibinfo{person}{E. Casett}, \bibinfo{person}{J.~C.
  R{\'e}us}, \bibinfo{person}{J. Stuginski-Barbosa}, \bibinfo{person}{A.
  Porporatti}, \bibinfo{person}{M. Carra}, \bibinfo{person}{M. Peres},
  \bibinfo{person}{G.~De~Luca Canto}, {and} \bibinfo{person}{D. Manfredini}.}
  \bibinfo{year}{2017}\natexlab{}.
\newblock \showarticletitle{Validity of different tools to assess sleep
  bruxism: a meta‐analysis}.
\newblock \bibinfo{journal}{\emph{Journal of Oral Rehabilitation}}
  \bibinfo{volume}{44} (\bibinfo{year}{2017}), \bibinfo{pages}{722–734}.
\newblock


\bibitem[\protect\citeauthoryear{Castroflorio, Bargellini, Rossini, Cugliari,
  Deregibus, and Manfredini}{Castroflorio et~al\mbox{.}}{2015}]%
        {emg-accuracy}
\bibfield{author}{\bibinfo{person}{T. Castroflorio}, \bibinfo{person}{A.
  Bargellini}, \bibinfo{person}{G. Rossini}, \bibinfo{person}{G. Cugliari},
  \bibinfo{person}{A. Deregibus}, {and} \bibinfo{person}{D. Manfredini}.}
  \bibinfo{year}{2015}\natexlab{}.
\newblock \showarticletitle{Agreement between clinical and portable EMG/ECG
  diagnosis of sleep bruxism}.
\newblock \bibinfo{journal}{\emph{Journal of Oral Rehabilitation}}
  \bibinfo{volume}{42}, \bibinfo{number}{10} (\bibinfo{year}{2015}),
  \bibinfo{pages}{759--764}.
\newblock


\bibitem[\protect\citeauthoryear{Castroflorio, Mesin, Tartaglia, Sforza, and
  Farina}{Castroflorio et~al\mbox{.}}{2013}]%
        {SBNaturalEnviorment:online}
\bibfield{author}{\bibinfo{person}{Tommaso Castroflorio}, \bibinfo{person}{Luca
  Mesin}, \bibinfo{person}{Gianluca~Martino Tartaglia},
  \bibinfo{person}{Chiarella Sforza}, {and} \bibinfo{person}{Dario Farina}.}
  \bibinfo{year}{2013}\natexlab{}.
\newblock \showarticletitle{Use of Electromyographic and Electrocardiographic
  Signals to Detect Sleep Bruxism Episodes in a Natural Environment}.
\newblock \bibinfo{journal}{\emph{IEEE Journal of Biomedical and Health
  Informatics}} \bibinfo{volume}{17}, \bibinfo{number}{6}
  (\bibinfo{year}{2013}), \bibinfo{pages}{994--1001}.
\newblock
\urldef\tempurl%
\url{https://doi.org/10.1109/JBHI.2013.2274532}
\showDOI{\tempurl}


\bibitem[\protect\citeauthoryear{{Dadnam D. and Dadnam C. and Al-Saffar
  H.}}{{Dadnam D. and Dadnam C. and Al-Saffar H.}}{2021}]%
        {PandemicAndBruxism:online}
\bibfield{author}{\bibinfo{person}{{Dadnam D. and Dadnam C. and Al-Saffar H.}}}
  \bibinfo{year}{2021}\natexlab{}.
\newblock \bibinfo{title}{{Pandemic Bruxism | British Dental Journal 230,
  271}}.
\newblock
  \bibinfo{howpublished}{\url{https://www.nature.com/articles/s41415-021-2788-3}}.
\newblock


\bibitem[\protect\citeauthoryear{González and Lantada}{González and
  Lantada}{2009}]%
        {forceSensor:online}
\bibfield{author}{\bibinfo{person}{C. González} {and} \bibinfo{person}{A~Díaz
  Lantada}.} \bibinfo{year}{2009}\natexlab{}.
\newblock \showarticletitle{A wearable passive force sensor powered by an
  active interrogator intended for intra-splint use for the detection and
  recording of bruxism}. In \bibinfo{booktitle}{\emph{2009 3rd International
  Conference on Pervasive Computing Technologies for Healthcare}}.
  \bibinfo{pages}{1--4}.
\newblock
\urldef\tempurl%
\url{https://doi.org/10.4108/ICST.PERVASIVEHEALTH2009.5884}
\showDOI{\tempurl}


\bibitem[\protect\citeauthoryear{Jirakittayakorn and Wongsawat}{Jirakittayakorn
  and Wongsawat}{2014}]%
        {emg-limitations}
\bibfield{author}{\bibinfo{person}{Nantawachara Jirakittayakorn} {and}
  \bibinfo{person}{Yodchanan Wongsawat}.} \bibinfo{year}{2014}\natexlab{}.
\newblock \showarticletitle{An EMG instrument designed for bruxism detection on
  masseter muscle}. In \bibinfo{booktitle}{\emph{The 7th 2014 Biomedical
  Engineering International Conference}}. \bibinfo{pages}{1--5}.
\newblock


\bibitem[\protect\citeauthoryear{Kawsar, Min, Mathur, and Montanari}{Kawsar
  et~al\mbox{.}}{2018}]%
        {kawsar-2018-esense}
\bibfield{author}{\bibinfo{person}{Fahim Kawsar}, \bibinfo{person}{Chulhong
  Min}, \bibinfo{person}{Akhil Mathur}, {and} \bibinfo{person}{Alessandro
  Montanari}.} \bibinfo{year}{2018}\natexlab{}.
\newblock \showarticletitle{Earables for Personal-Scale Behavior Analytics}.
\newblock \bibinfo{journal}{\emph{IEEE Pervasive Computing}}
  \bibinfo{volume}{17}, \bibinfo{number}{3} (\bibinfo{year}{2018}),
  \bibinfo{pages}{83--89}.
\newblock
\urldef\tempurl%
\url{https://doi.org/10.1109/MPRV.2018.03367740}
\showDOI{\tempurl}


\bibitem[\protect\citeauthoryear{Khanna, Srivastava, Pan, Jain, and
  Nguyen}{Khanna et~al\mbox{.}}{2021}]%
        {JawSense6:online}
\bibfield{author}{\bibinfo{person}{Prerna Khanna}, \bibinfo{person}{Tanmay
  Srivastava}, \bibinfo{person}{Shijia Pan}, \bibinfo{person}{Shubham Jain},
  {and} \bibinfo{person}{Phuc Nguyen}.} \bibinfo{year}{2021}\natexlab{}.
\newblock \showarticletitle{JawSense: Recognizing Unvoiced Sound Using a
  Low-Cost Ear-Worn System}. In \bibinfo{booktitle}{\emph{Proceedings of the
  22nd International Workshop on Mobile Computing Systems and Applications}}
  (Virtual, United Kingdom) \emph{(\bibinfo{series}{HotMobile '21})}.
  \bibinfo{publisher}{Association for Computing Machinery},
  \bibinfo{pages}{44–49}.
\newblock
\urldef\tempurl%
\url{https://doi.org/10.1145/3446382.3448363}
\showDOI{\tempurl}


\bibitem[\protect\citeauthoryear{Kinjo, Wada, Churei, Ohmi, Hayashi, Yagishita,
  Uo, and Ueno}{Kinjo et~al\mbox{.}}{2021}]%
        {mouthGuard:online}
\bibfield{author}{\bibinfo{person}{Rio Kinjo}, \bibinfo{person}{Takahiro Wada},
  \bibinfo{person}{Hiroshi Churei}, \bibinfo{person}{Takehiro Ohmi},
  \bibinfo{person}{Kairi Hayashi}, \bibinfo{person}{Kazuyoshi Yagishita},
  \bibinfo{person}{Motohiro Uo}, {and} \bibinfo{person}{Toshiaki Ueno}.}
  \bibinfo{year}{2021}\natexlab{}.
\newblock \showarticletitle{Development of a Wearable Mouth Guard Device for
  Monitoring Teeth Clenching during Exercise}.
\newblock \bibinfo{journal}{\emph{Sensors}} \bibinfo{volume}{21},
  \bibinfo{number}{4} (\bibinfo{year}{2021}).
\newblock
\urldef\tempurl%
\url{https://doi.org/10.3390/s21041503}
\showDOI{\tempurl}


\bibitem[\protect\citeauthoryear{Knierim, Schemmer, and Woehler}{Knierim
  et~al\mbox{.}}{2021}]%
        {aroundEar:online}
\bibfield{author}{\bibinfo{person}{Michael Knierim}, \bibinfo{person}{Max
  Schemmer}, {and} \bibinfo{person}{Dominik Woehler}.}
  \bibinfo{year}{2021}\natexlab{}.
\newblock \showarticletitle{Detecting Daytime Bruxism Through Convenient and
  Wearable Around-the-Ear Electrodes}. In \bibinfo{booktitle}{\emph{12th
  International Conference on Applied Human Factors and Ergonomics (AHFE
  2021)}}.
\newblock


\bibitem[\protect\citeauthoryear{Ma, Ferlini, and Mascolo}{Ma
  et~al\mbox{.}}{2021}]%
        {dong-2021}
\bibfield{author}{\bibinfo{person}{Dong Ma}, \bibinfo{person}{Andrea Ferlini},
  {and} \bibinfo{person}{Cecilia Mascolo}.} \bibinfo{year}{2021}\natexlab{}.
\newblock \showarticletitle{OESense: Employing Occlusion Effect for in-Ear
  Human Sensing}. In \bibinfo{booktitle}{\emph{Proceedings of the 19th Annual
  International Conference on Mobile Systems, Applications, and Services}}
  (Virtual Event, Wisconsin) \emph{(\bibinfo{series}{MobiSys '21})}.
  \bibinfo{publisher}{Association for Computing Machinery},
  \bibinfo{pages}{175–187}.
\newblock
\urldef\tempurl%
\url{https://doi.org/10.1145/3458864.3467680}
\showDOI{\tempurl}


\bibitem[\protect\citeauthoryear{Min, Mathur, and Kawsar}{Min
  et~al\mbox{.}}{2018}]%
        {eSense2:online}
\bibfield{author}{\bibinfo{person}{Chulhong Min}, \bibinfo{person}{Akhil
  Mathur}, {and} \bibinfo{person}{Fahim Kawsar}.}
  \bibinfo{year}{2018}\natexlab{}.
\newblock \showarticletitle{Exploring Audio and Kinetic Sensing on Earable
  Devices}. In \bibinfo{booktitle}{\emph{Proceedings of the 4th ACM Workshop on
  Wearable Systems and Applications}} (Munich, Germany)
  \emph{(\bibinfo{series}{WearSys '18})}. \bibinfo{publisher}{Association for
  Computing Machinery}, \bibinfo{pages}{5–10}.
\newblock
\urldef\tempurl%
\url{https://doi.org/10.1145/3211960.3211970}
\showDOI{\tempurl}


\bibitem[\protect\citeauthoryear{Pigozzi, Rehm, Fagondes, Pellizzer, and
  Grossi}{Pigozzi et~al\mbox{.}}{2019}]%
        {currentMethods:online}
\bibfield{author}{\bibinfo{person}{Lucas~B Pigozzi}, \bibinfo{person}{Daniela
  D~S Rehm}, \bibinfo{person}{Simone~C Fagondes}, \bibinfo{person}{Eduardo~P
  Pellizzer}, {and} \bibinfo{person}{Márcio~L Grossi}.}
  \bibinfo{year}{2019}\natexlab{}.
\newblock \showarticletitle{Current Methods of Bruxism Diagnosis: A Short
  Communication}.
\newblock \bibinfo{journal}{\emph{The International journal of prosthodontics}}
  \bibinfo{volume}{32}, \bibinfo{number}{3} (\bibinfo{year}{2019}),
  \bibinfo{pages}{263—264}.
\newblock
\urldef\tempurl%
\url{https://doi.org/10.11607/ijp.6196}
\showDOI{\tempurl}


\bibitem[\protect\citeauthoryear{Prakash, Yang, Wei, Hassanieh, and
  Choudhury}{Prakash et~al\mbox{.}}{2020}]%
        {EarSense48:online}
\bibfield{author}{\bibinfo{person}{Jay Prakash}, \bibinfo{person}{Zhijian
  Yang}, \bibinfo{person}{Yu-Lin Wei}, \bibinfo{person}{Haitham Hassanieh},
  {and} \bibinfo{person}{Romit~Roy Choudhury}.}
  \bibinfo{year}{2020}\natexlab{}.
\newblock \showarticletitle{EarSense: Earphones as a Teeth Activity Sensor}. In
  \bibinfo{booktitle}{\emph{Proceedings of the 26th Annual International
  Conference on Mobile Computing and Networking}} (London, United Kingdom)
  \emph{(\bibinfo{series}{MobiCom '20})}. \bibinfo{publisher}{Association for
  Computing Machinery}, Article \bibinfo{articleno}{40}.
\newblock
\urldef\tempurl%
\url{https://doi.org/10.1145/3372224.3419197}
\showDOI{\tempurl}


\bibitem[\protect\citeauthoryear{Rupavatharam and Gruteser}{Rupavatharam and
  Gruteser}{2019}]%
        {inEarClenching:online}
\bibfield{author}{\bibinfo{person}{Siddharth Rupavatharam} {and}
  \bibinfo{person}{Marco Gruteser}.} \bibinfo{year}{2019}\natexlab{}.
\newblock \showarticletitle{Towards In-Ear Inertial Jaw Clenching Detection}.
  In \bibinfo{booktitle}{\emph{Proceedings of the 1st International Workshop on
  Earable Computing}} (London, United Kingdom)
  \emph{(\bibinfo{series}{EarComp'19})}. \bibinfo{publisher}{Association for
  Computing Machinery}, \bibinfo{pages}{54–55}.
\newblock
\urldef\tempurl%
\url{https://doi.org/10.1145/3345615.3361134}
\showDOI{\tempurl}


\bibitem[\protect\citeauthoryear{{Shetty, S., Pitti, V., C.L., Babu, S., G.P.,
  Kumar, S. and Deepthi, B. C. }}{{Shetty, S., Pitti, V., C.L., Babu, S., G.P.,
  Kumar, S. and Deepthi, B. C. }}{2010}]%
        {BruxismALitratureReview:online}
\bibfield{author}{\bibinfo{person}{{Shetty, S., Pitti, V., C.L., Babu, S.,
  G.P., Kumar, S. and Deepthi, B. C. }}.} \bibinfo{year}{2010}\natexlab{}.
\newblock \bibinfo{title}{{ Bruxism: A Literature Review. J Indian Prosthodont
  Soc 10, 141–148 (2010).}}
\newblock
  \bibinfo{howpublished}{\url{https://link.springer.com/article/10.1007/s13191-011-0041-5}}.
\newblock


\bibitem[\protect\citeauthoryear{van Dijk}{van Dijk}{1995}]%
        {biteGuard:online}
\bibfield{author}{\bibinfo{person}{M. van Dijk}.}
  \bibinfo{year}{1995}\natexlab{}.
\newblock \showarticletitle{The binary symmetric broadcast channel with
  confidential messages, with tampering}. In
  \bibinfo{booktitle}{\emph{Proceedings of 1995 IEEE International Symposium on
  Information Theory}}. \bibinfo{pages}{487--}.
\newblock
\urldef\tempurl%
\url{https://doi.org/10.1109/ISIT.1995.550474}
\showDOI{\tempurl}


\bibitem[\protect\citeauthoryear{Yachida, Arima, Castrillon, Baad-Hansen,
  Ohata, and Svensson}{Yachida et~al\mbox{.}}{2016}]%
        {validitySelfReport:online}
\bibfield{author}{\bibinfo{person}{Wataru Yachida}, \bibinfo{person}{Taro
  Arima}, \bibinfo{person}{Eduardo~E. Castrillon}, \bibinfo{person}{Lene
  Baad-Hansen}, \bibinfo{person}{Noboru Ohata}, {and} \bibinfo{person}{Peter
  Svensson}.} \bibinfo{year}{2016}\natexlab{}.
\newblock \showarticletitle{Diagnostic validity of self-reported measures of
  sleep bruxism using an ambulatory single-channel EMG device}.
\newblock \bibinfo{journal}{\emph{Journal of Prosthodontic Research}}
  \bibinfo{volume}{60}, \bibinfo{number}{4} (\bibinfo{year}{2016}),
  \bibinfo{pages}{250--257}.
\newblock
\showISSN{1883-1958}
\urldef\tempurl%
\url{https://doi.org/10.1016/j.jpor.2016.01.001}
\showDOI{\tempurl}


\bibitem[\protect\citeauthoryear{{Zoom Video Communications Inc.}}{{Zoom Video
  Communications Inc.}}{2016}]%
        {ZoomSecure:online}
\bibfield{author}{\bibinfo{person}{{Zoom Video Communications Inc.}}}
  \bibinfo{year}{2016}\natexlab{}.
\newblock \bibinfo{title}{{Zoom Video Communications Inc . (2016). Security
  guide. Zoom Video Communications Inc.}}
\newblock
  \bibinfo{howpublished}{\url{https://d24cgw3uvb9a9h.cloudfront.net/static/81625/doc/Zoom-Security-White-Paper.pdf}}.
\newblock


\end{thebibliography}

\end{document}